\documentclass[conference]{IEEEtran}
\IEEEoverridecommandlockouts
\usepackage{cite}
\usepackage{amsmath,amssymb,amsfonts}
\usepackage{algorithmic}
\usepackage{graphicx}
\usepackage{textcomp}
\usepackage{xcolor}
\def\BibTeX{{\rm B\kern-.05em{\sc i\kern-.025em b}\kern-.08em
    T\kern-.1667em\lower.7ex\hbox{E}\kern-.125emX}}

\usepackage{array}
\newcolumntype{?}{!{\vrule width 2pt}}

\usepackage{hyperref}
\definecolor{urlcolor} {rgb}{0,0,0.9333333333333}
\hypersetup{
    colorlinks=true,     
    urlcolor=urlcolor,
    linkcolor=black,
    filecolor=black,
    citecolor=black,
    menucolor=black,
    runcolor=black
    }

\makeatletter
\newcommand\footnoteref[1]{\protected@xdef\@thefnmark{\ref{#1}}\@footnotemark}
\makeatother

\makeatletter 
\newcommand{\linebreakand}{%
  \end{@IEEEauthorhalign}
  \hfill\mbox{}\par
  \mbox{}\hfill\begin{@IEEEauthorhalign}
}
\makeatother 
    
\begin{document}

\title{Structure-Preserving Transformers for Sequences of SPD Matrices
\thanks{
{MS} is supported by the French National Research Agency (ANR) and Region Normandie under grant HAISCoDe. This work was granted access to the HPC resources of IDRIS under the allocation 2022-AD010613618 made by GENCI, and to the computing resources of CRIANN (Normandy, France). {This work was partially carried out during a CNRS leave at GREYC of FY.}}
}

\author{\IEEEauthorblockN{Mathieu Seraphim}
\IEEEauthorblockA{\textit{Normandie Univ, UNICAEN,} \\
\textit{ENSICAEN, CNRS, GREYC}\\
14000 Caen, France \\
mathieu.seraphim@unicaen.fr}
\and
\IEEEauthorblockN{Alexis Lechervy}
\IEEEauthorblockA{\textit{Normandie Univ, UNICAEN,} \\
\textit{ENSICAEN, CNRS, GREYC}\\
14000 Caen, France \\
alexis.lechervy@unicaen.fr
}
\and
\IEEEauthorblockN{Florian Yger}
\IEEEauthorblockA{\textit{LAMSADE, CNRS,}\\
\textit{PSL Univ. Paris-Dauphine}\\
75016 Paris, France \\
florian.yger@lamsade.dauphine.fr
}
\and
\linebreakand
\IEEEauthorblockN{Luc Brun}
\IEEEauthorblockA{\textit{Normandie Univ, ENSICAEN,} \\
\textit{UNICAEN, CNRS, GREYC}\\
14000 Caen, France \\
luc.brun@ensicaen.fr
}
\and
\IEEEauthorblockN{Olivier Etard}
\IEEEauthorblockA{\textit{Normandie Université, UNICAEN, INSERM,} \\
\textit{COMETE, CYCERON, CHU Caen}\\
14000 Caen, France \\
olivier.etard@unicaen.fr
}
}

\maketitle

\begin{abstract}
In recent years, Transformer-based auto-attention mechanisms have been successfully applied to the analysis of a variety of context-reliant data types, from texts to images and beyond, including data from non-Euclidean geometries. In this paper, we present such a mechanism, designed to classify sequences of Symmetric Positive Definite matrices while preserving their Riemannian geometry throughout the analysis. We apply our method to automatic sleep staging on timeseries of EEG-derived covariance matrices from a standard dataset, obtaining high levels of stage-wise performance.
\end{abstract}


\begin{figure}
  \begin{center}
  \includegraphics[width=.4\textwidth]{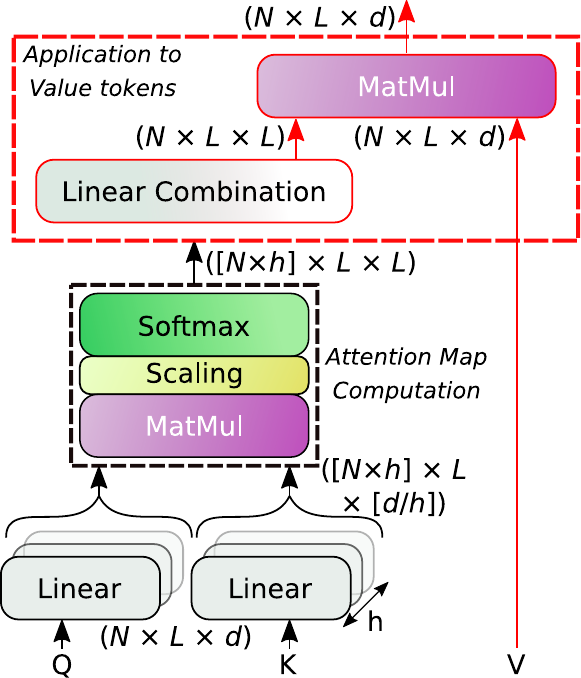}
  \caption{The SP-MHA architecture. In parentheses are tensor dimensions at every step, with $N$ the batch size.}
  \label{fig:SPMHA}
  \end{center}
\end{figure}

\section{Introduction}
\label{sec:intro}

When analyzing the relationship between concurrent signals, covariance matrices are a useful tool, with applications in fields like Brain-Computer Interfaces (BCI)~\cite{BCI1} and evolutionary computation~\cite{cma_es}.
By construction, they are rich in information, illustrating the relationship between signals while still encoding for signal-wise information on their diagonal.
Such matrices are at least Positive Semi-Definite, and often fully Symmetric Positive Definite (SPD).
The set of $n \times n$ SPD matrices ($SPD(n)$) is a non-Euclidean, Riemannian (i.e. metric) manifold, and the regular Euclidean operations of most Neural Network (NN)-based models seldom preserve that geometric structure, introducing deformations such as the ``swelling effect"~\cite{LogEuclidean}.
Structure-preserving NN-based approaches have been introduced~\cite{huang2017spdnet,manifoldnet}, deriving their layers from one of two geodesic-defining metrics on $SPD(n)$. Affine invariant metrics offer the best properties, but present computational challenges (e.g. no closed-form formula for averaging)~\cite{affine_invariant}.
LogEuclidean metrics are less isotropic, but still prevent swelling while being easier to compute~\cite{LogEuclidean}.

\begin{figure*}
  \begin{center}
  \includegraphics[width=.9\textwidth]{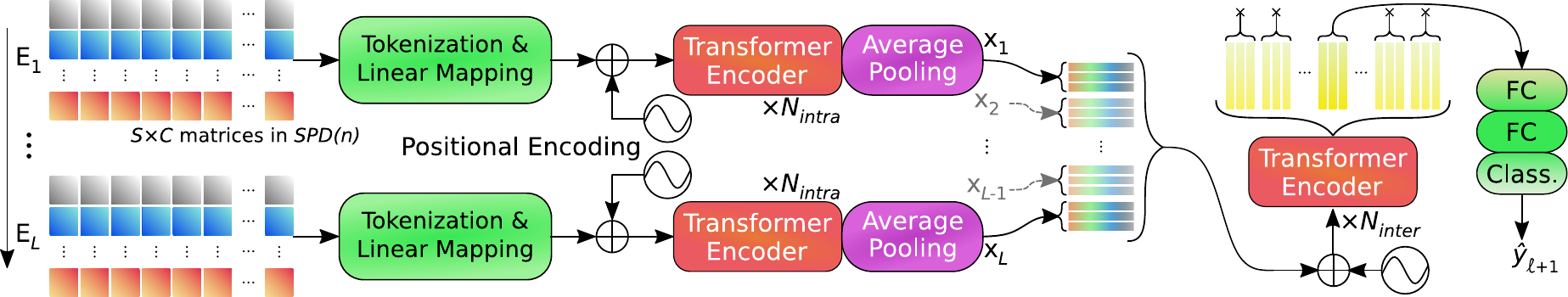}
  \caption{SPDTransNet global architecture, with $t=3$ feature tokens per epoch.}
  \label{fig:architecture}
  \end{center}
\end{figure*}

In this paper, we present a structure-preserving self-attention mechanism applicable to sequences of SPD matrices, derived from {such a} LogEuclidean metric. We embed said mechanism into a Transformer-based architecture, and apply it to a biomedical classification problem.
Transformer-based technology has exploded in popularity ever since its introduction in~\cite{transformers}, with self-attention mechanisms being applied to very different problems.
With regards to Riemannian geometry, innovations seem centered around the computation and application of attention maps, specifically. For instance, Konstantinidis et al.~\cite{Konstantinidis2022MultimanifoldAF} combine the standard attention maps with Grassmann and SPD manifold-valued maps, to enrich their computer vision model's descriptive capabilities. By contrast, both He et al.~\cite{he_gauge_equivariant} and Li et al.~\cite{li2022geodesic} developed architectures to analyze 2D-manifold-valued data in 3D space, the former achieving rotational equivariance with respect to surfaces on the manifold and the latter developing two geodesic distances applicable to point clouds, and building attention maps from these distances. More generally, Kratsios et al.~\cite{kratsios2022universal} provide a mathematical framework to apply attention mechanisms on a variety of constrained sets, including manifolds.
While the latter approaches share our interest in preserving geometric information, little to no focus is given to a Transformer's other components.
{Although simple single-head attention modules for SPD-valued data have been recently developed~\cite{pan2022,qin2024}, t}o the best of our knowledge, our approach is the only one {utilizing full structure-preserving Transformer encoders in this context}.

\section{SPD Structure-Preserving Attention}
\label{sec:theory}

{The LogEuclidean distance (Section~\ref{sec:intro}) can be written as:}
\begin{equation}
\label{eq:LE}
\delta_{LE}(A, B) = \lVert log_{mat}(A) - log_{mat}(B) \rVert_2
\end{equation}
{with $A$, $B$ $\in$ $SPD(n)$.}
Here, $\lVert X \rVert_2$ ({with} $X$ $\in$ $Sym(n)$) is the $\mathcal{L}_2$ norm applied to the upper triangular of $X$, {and $log_{mat}(\cdot)$ is the matrix logarithm}, bijectively mapping $SPD(n)$ onto $Sym(n)$, the vector space of $n \times n$ symmetric matrices (with $exp_{mat}(\cdot)$ being its inverse).
{Euclidean operations on $Sym(n)$ are thus equivalent to LogEuclidean (and therefore Riemannian) operations on the corresponding SPD matrices.}

Let $B_{n} = \{e_{i, j}\}_{0 < i \leq j} \subset \mathbb{R}^{n \times n}$ be the the canonical basis of $Sym(n)$, with $(e_{i, j})_{i, j} = (e_{i, j})_{j, i} = 1$, and all other coefficients at 0. Let the triangular number $d(n) = \frac{n(n+1)}{2}$ be the dimension of $Sym(n)$.
Any matrix $M$ of $Sym(n)$ can be written in the basis $B_{n}$ as a vector of coordinates in $\mathbb{R}^{d(n)}$.

{In accordance with convention surrounding Transformer-based architectures, we refer to these vectors as ``tokens".
In this paper, any token of $\mathbb{R}^{d(n)}$ is thus equivalent to a matrix in $SPD(n)$, and linear combinations of such tokens would equate to a LogEuclidean weighted sum in $SPD(n)$, preserving their underlying manifold structure.}

\subsection{Structure-Preserving Multihead Attention (SP-MHA)}
\label{ssec:SPMHA}

In the original Linear Multihead Attention (L-MHA) component of Transformers~\cite{transformers}, the input tokens in the Q, K and V tensors are processed in parallel in $h$ attention heads, then recombined through concatenation.
There is no guarantee that any underlying SPD structure in our tokens would survive this concatenation. Echoing the similar concerns, Li et al.~\cite{li2022geodesic} decided to forego having multiple heads.
{Likewise, Pan et al.~\cite{pan2022} and Qin et al.~\cite{qin2024} restricted themselves to single-head SPD-valued attention modules.
By contrast, we design our Multihead Attention block to retain} the parallel attention maps computation {of the original L-MHA} without sacrificing our data's structure.

Let $d(m)$ be the dimension of input tokens.
As seen in Figure~\ref{fig:SPMHA}, our SP-MHA block does the following:
\begin{equation}
\label{eq:SPMHA}
MHA_{SP}(Q, K, V) = C\left(sm\left(\dfrac{\mathcal{L}_Q(Q) \cdot \mathcal{L}_K(K)^T}{\sqrt{d(m)/h}}\right)\right) \cdot V
\end{equation}
with $\mathcal{L}_Q(\cdot)$ and $\mathcal{L}_K(\cdot)$ banks of $h$ linear maps from $\mathbb{R}^{d(m)}$ to $\mathbb{R}^{\frac{d(m)}{h}}$, $sm(\cdot)$ the softmax function, and $C(\cdot)$ the weighted linear combination of the $h$ post-softmax attention maps.
Here, the computation of attention maps (small-dashed black rectangle in Figure~\ref{fig:SPMHA}) remains identical to L-MHA. However, their application to V (large-dashed red rectangle in the figure) only requires a matrix multiplication, i.e. linear combinations of V's tokens weighted by the combined attention map.
As such, the SP-MHA block does not compromise our tokens' vector space geometry.

\subsection{Triangular linear maps}
\label{ssec:justification}

Let $Sym(n)$ and $Sym(m)$ have the canonical bases $B_{n}$ and $B_{m}$, respectively. Let $\mathcal{L}_{n, m}(\cdot)$ be a linear map from $Sym(n)$ to $Sym(m)$, represented by the matrix $W$ in $\mathbb{R}^{d(m) \times d(n)}$ with respect to the bases (implemented in code through a fully connected NN layer between tokenized matrices). We shall refer to such a map as a ``triangular" linear map.

Let $A^*, B^*$ be in $SPD(n)$, mapped to $A, B \in Sym(n)$ through $log_{mat}(\cdot)$. As $\mathcal{L}_{n, m}(\cdot)$ is a continuous linear map:
\begin{equation}
\label{eq:ineq_1}
\lVert \mathcal{L}_{n, m}(A) - \mathcal{L}_{n, m}(B) \rVert_2 \leq \lVert W \rVert_* \cdot \lVert A - B \rVert_2
\end{equation}
\begin{equation}
\label{eq:ineq_2}
\delta_{LE}(\mathcal{L}_{n, m}^\mathcal{R}(A^*), \mathcal{L}_{n, m}^\mathcal{R}(B^*)) \leq \lVert W \rVert_* \cdot \delta_{LE}(A^*, B^*)
\end{equation}
with $\lVert \cdot \rVert_*$ the matrix norm induced by the norm $\lVert \cdot \rVert_2$, and $\mathcal{L}_{n, m}^\mathcal{R}(\cdot) = exp_{mat} \circ \mathcal{L}_{n, m} \circ log_{mat}(\cdot)$ mapping $SPD(n)$ onto $SPD(m)$.
By definition of $\delta_{LE}$ (Equation~\ref{eq:LE}), Equations~\ref{eq:ineq_1} and~\ref{eq:ineq_2} are strictly identical.
Hence, applying $\mathcal{L}_{n, m}(\cdot)$ on our tokens is equivalent to applying $\mathcal{L}_{n, m}^\mathcal{R}(\cdot)$ on matrices in $SPD(n)$. The output tokens exhibit the Riemannian structure of $SPD(m)$, and relations of proximity are preserved. Therefore, so is the overall structure of our data.

Note that while other SPD-to-SPD NN-based mappings have been proposed~\cite{SPD_dim_reduc,huang2017spdnet}, they rely on full-rank weight tensors, whereas $\mathcal{L}_{n, m}^\mathcal{R}(\cdot)$ does not require special constraints.

\section{Application to EEG Sleep Staging}
\label{sec:EEG}

The study of sleep most often requires the analysis of electrophysiological - including electroencephalographic (EEG) - signals, subdivided into fixed-length windows (``epochs") and manually labeled with the appropriate sleep stages, inferred from properties of the signal in and around each epoch~\cite{berry2017aasm}.

As seen in a recent survey by Phan et al.~\cite{Phan_survey}, state-of-the-art automatic sleep staging models typically use two-step architectures -  given a sequence of epochs, epoch-wise features are extracted before being compared at the sequence-wise level, utilizing this contextual information to improve classification.
Since epochs often contain markers indicative of multiple stages, two-step architectures tend to subdivide them further, extracting features from subwindows using convolutional NNs~\cite{Supratak2017,SEO2020102037} and/or recurrent NNs~\cite{phan2022xsleepnet,phan2023lseqsleepnet,guillot2021} - the latter utilizing RNNs for both steps.
Multiple authors have adapted this context-inclusive approach to Transformer-based architectures~\cite{zhu2020,phan2022sleeptransformer,CAIP_article}, with auto-attention mechanisms at both the intra- and inter-epoch levels, taking advantage of the high performance they offer when applied to sequence-based data.

\subsection{{The stakes of automatic sleep staging}}
\label{ssec:N1_performance}

\begin{table*}[ht]
\small
\centering
\begin{tabular}{ |c|c?c|c|c?c|c|c| }
    \cline{2-8}
    \multicolumn{1}{c|}{} & Model & MF1 & Macro Acc. & N1 F1 & Valid. metric & Token dim. $d(m)$ & \# Feat. Tokens $t$ \\\hline
    1 & SPDTransNet, $L=13$ & 81.06 $\pm$ 3.49 & \textbf{84.87} $\pm$ 2.47 & 60.39 $\pm$ 6.77 & MF1 & 351 ($m = 26$) & 7 \\\hline
    2 & SPDTransNet, $L=21$ & \textbf{81.24} $\pm$ 3.29 & 84.40 $\pm$ 2.61 & \textbf{60.50} $\pm$ 6.18 & MF1 & 351 ($m = 26$) & 10 \\\hline
    3 & SPDTransNet, $L=29$ & 80.83 $\pm$ 3.40 & 84.29 $\pm$ 2.65 & 60.35 $\pm$ 6.01 & N1 F1 &  351 ($m = 26$) & 5 \\\hline
    4 & Classic MHA & 80.82 $\pm$ 3.40 & 84.60 $\pm$ 2.95 & 60.16 $\pm$ 7.20 & MF1 &  351 ($m = 26$) & 10 \\\hline
    5 & DeepSleepNet~\cite{Supratak2017} & 78.14 $\pm$ 4.12 & 80.05 $\pm$ 3.47 & 53.52 $\pm$ 8.24 & N/A & N/A & N/A \\\hline
    6 & IITNet~\cite{SEO2020102037} & 78.48 $\pm$ 3.15 & 81.88 $\pm$ 2.89 & 56.01 $\pm$ 6.54 & N/A & N/A & N/A \\\hline
    7 & GraphSleepNet~\cite{jia2020graphsleepnet} & 75.58 $\pm$ 3.75 & 79.75 $\pm$ 3.41 & 50.80 $\pm$ 8.06 & N/A & N/A & N/A \\\hline
    8 & Dequidt et al.~\cite{paul} & 81.04 $\pm$ 3.26 & 82.59 $\pm$ 3.45 & 58.42 $\pm$ 6.09 & N/A & N/A & N/A \\\hline
    9 & Seraphim et al.~\cite{CAIP_article} & 79.78 $\pm$ 4.56 & 81.76 $\pm$ 4.61 & 58.43 $\pm$ 6.41 & MF1 & Concatenation & 1 \\\hline
\end{tabular}
\caption{Results obtained from both our model and the re-trained literature. Best results are in \textbf{bold}.}
\label{tab:results}
\end{table*}

According to the aforementioned survey~\cite{Phan_survey}, current sleep staging models have attained a sufficient performance level to replace manual staging in some contexts. However, we have found that class-wise performance was often lacking, particularly with regards to the N1 sleep stage~\cite{berry2017aasm}, universally difficult to classify.
Most EEG datasets are heavily imbalanced, with the N1 stage often underrepresented (Section~\ref{sec:experiments}) - models optimized for high overall accuracy may thus sacrifice N1 classification if it improves global performance.
To account for this, recent approaches~\cite{paul,CAIP_article} elected to primarily evaluate their performance through the macro-averaged F1 (MF1) score, a class-wise balanced metric widely used in the literature. They also rebalance their training sets through oversampling, so that all stages within have the same number of classification targets.
While the survey states that a sequence-to-sequence classification scheme (classifying each epoch in the input sequence) might lead to better performance, having multilabel inputs is nonsensical for this rebalancing - hence their use of a sequence-to-epoch scheme (classifying one epoch per sequence).

Beyond sleep staging, EEG signals are also utilized in BCI (Section~\ref{sec:intro}), where they are often analyzed through the lens of functional connectivity - the activation correlations between different brain regions~\cite{EICKHOFF2015187}. Automatic sleep staging through functional connectivity was first investigated by Jia et al.~\cite{jia2020graphsleepnet}, using epoch-wise graph learning to estimate said connectivity and sequence-wise spatio-temporal graph NNs to compare them.
By contrast, Seraphim et al.~\cite{CAIP_article} estimate it through covariance matrices, as is commonly done in BCI~\cite{BCI1}.
Their two-step model uses standard Transformer encoders at each step, reminiscent of~\cite{phan2022sleeptransformer}. Each input epoch is described as a multichannel timeseries of SPD matrices, which are then tokenized bijectively.
However, their approach does not guarantee the preservation of their data's SPD structure, as they operate a channel-wise concatenation of their tokens, in addition to the concatenations found within their encoders (Section~\ref{ssec:SPMHA}).
Hence, we propose a Transformer-based model capable of analyzing EEG-derived functional connectivity through SPD matrices \textit{without} sacrificing the SPD structure of our data throughout the analysis.

\subsection{{Our} preprocessing}
\label{ssec:preprocessing}

To estimate functional connectivity from {EEG} signals, we apply the same preprocessing pipeline as~\cite{CAIP_article}\footnote{\label{fn:Git}More details at github.com/MathieuSeraphim/SPDTransNet.}.
{We first select $n$ EEG signals. Each signal is then filtered along $C$ frequency bands, divided into epochs, and further subdivided into $S$ subwindows per epoch.
A covariance matrix is computed per channel and subwindow, resulting in $S \times C$ covariance matrices in $SPD(n)$ for each epoch.}
We then augment our matrices with signal-derived information before whitening them\footnoteref{fn:Git}, leading to more uniformly distributed matrices in $SPD(n+1)$.
Said whitening requires the computation of average covariance matrices per recording and channel, which was done in~\cite{CAIP_article} by computing the covariances over the entire recording. Instead, we average all relevant matrices using the standard affine invariant metric~\cite{affine_invariant}, improving performance.

\subsection{The SPDTransNet model}
\label{ssec:model}

As can be seen in Figure~\ref{fig:architecture}, our SPDTransNet model takes as input a sequence of $L$ epochs, composed of a central epoch to classify and surrounding epochs to provide context.
Given $\ell$ the context size, we have $L = 2 \cdot \ell + 1$.

{Our preprocessing yields $S \times C$ matrices of $SPD(n+1)$ per epoch (Section~\ref{ssec:preprocessing}).}
Each {of these matrices} is mapped onto {$Sym(n+1)$} {through $log_{mat}(\cdot)$ and} tokenized (Section~\ref{sec:theory}). {Each input token of $\mathbb{R}^{d(n+1)}$ thus encodes the covariance of each signal pair, along with signal-specific information (the variance and augmentation features).}

These tokens are linearly mapped onto {$\mathbb{R}^{d(m)}$} (with $m>n+1$, as we have found that larger tokens improve performance).
The $S \times C$ grid of tokens is then arranged into a sequence, with the $S$ tokens in the channel 1 followed by the $S$ tokens in channel 2, etc.

At the intra-epoch level, a first positional encoding is applied to the tokens, which pass through the first Transformer encoder. The $S \times C$ output tokens are then uniformly divided into $t$ groups, with each group averaged into a single {feature} token.
The $L$ sets of $t$ {epoch-wise feature} tokens are then regrouped at the inter-epoch level, and passed through another positional encoding and Transformer encoder pair.
Finally, the {feature} tokens corresponding to the central epoch (of index $\ell + 1$ in Figure~\ref{fig:architecture}) go through two FC blocks (fully connected layers followed by ReLU activation and a dropout layer), and are mapped onto $\hat{y}_{\ell+1} \in \mathbb{R}^c$ by a final classification linear map, with $c$ the number of classes.

We ensure structure preservation by using the SP-MHA block in all Transformer encoders, and choosing all linear maps within said encoders' Feed-Forward (FF) components~\cite{transformers} to be triangular (Section~\ref{ssec:justification}).
The ReLU and dropout layers in the FF blocks do not cause issue, as setting a values within a token to 0 won't remove the corresponding matrix from $Sym(m)$.
Same for the positional encodings, average poolings and in-encoder layer normalizations, which all qualify as linear combinations.

As such, our model preserves the SPD structure of its input up to the final classification {layers, and every token throughout the model remains equivalent to an SPD matrix obtained through Riemannian operations (Section~\ref{sec:theory})}.

\section{Experiments \& Results}
\label{sec:experiments}

We utilize the MASS SS3 dataset~\cite{o2014montreal} due to its large number of available EEG electrode-derived signals and widespread use in the literature. It is composed of 62 full-night recordings of healthy subjects, segmented into 30s epochs. Due to its nature, it is unbalanced, with the largest and smallest of its $c = 5$ classes (stages N2 and N1) composed of 50.24\% and 8.16\% of the dataset, respectively.
{Out of 20 available electrodes,} we selected the {$n$ =} 8 electrodes F3, F4, C3, C4, T3, T4, O1 and O2{, providing us with a good coverage of the brain while limiting redundancies}.
{As in~\cite{CAIP_article}, we filter our signals to obtain $C$ = $7$ channels, and subdivide each epoch into $S$ = 30 one-second windows\footnoteref{fn:Git}, yielding us $30 \times 7$ matrices in $SPD(9)$ after preprocessing (Section~\ref{ssec:preprocessing}).}

To maximize class-wise performance, we operate a hyperparameter research per configuration, followed by a 31-fold cross-validation. As do~\cite{paul,CAIP_article} (Section~\ref{ssec:N1_performance}), we rebalance all training sets and maximize the MF1 score. 
To explore the importance of the context length $\ell$ (Section~\ref{ssec:model}) within our model, we ran hyperparameter researches with $\ell$ = 6, 10 or 14 (i.e. $L$ = 13, 21 or 29), with hyperparameter research configuration unchanged between them.

Our hyperparameter researches use the Optuna tool~\cite{akiba2019optuna}, with 5 simultaneous runs and 50 total runs per configuration. Hyperparameters include\footnoteref{fn:Git} the token size $d(m)$, set by the first linear map (Section~\ref{ssec:model}) and chosen in \{351, 378\} (i.e. $m$ $\in$ \{26, 27\})\footnote{\label{fn:d/h}Since $\frac{d(m)}{h}$ must be an integer, potential values for those are limited.};
the $h$ parameter of each Transformer encoder, in \{3, 9\}\footnoteref{fn:d/h};
and the number of epoch feature tokens $t$ (Section~\ref{ssec:model}), chosen among \{1, 3, 5, 7, 10\} - with in particular $t = 1$ akin to describing each epoch with a single token, and $t = 7$ corresponding to one token being preserved per channel.
We train all folds on the hyperparameters giving the best validation MF1, as well as those with the best F1 score for the N1 stage. Out of those two sets, the results from the set yielding the best average test MF1 is presented in lines 1 to 3 of Table~\ref{tab:results}, with the corresponding hyperparameter set, $d(m)$ and $t$ in the final three columns.

We obtain the best MF1 and N1 F1 scores for $L = 21$, whereas the best macro-averaged accuracy is obtained for $L = 13$.
For all values of $L$, we outperform the state-of-the-art on the considered metrics (except for the MF1 score for $L = 29$). Moreover, all three configurations have around a two-point lead in both macro accuracy and N1 F1 score.
While our model favors the smaller token size of $d(m) = 351$ for all values of $L$, it seems that having a large number of tokens to describe each epoch (at least $t = 5$) is necessary for best performance. 
Overall, $L = 21$ seems to be a good compromise to capture enough contextual information without burdening our model with irrelevant data.
We also investigate the impact of our strict structural preservation by replacing the SP-MHA block of SPDTransNet model with the classic L-MHA (Section~\ref{eq:SPMHA}), all other things being equal (with $L = 21$).
Results for this configuration are displayed in line 4 of the table.

We compare ourselves to five models:
DeepSleepNet~\cite{Supratak2017}, often used as a benchmark, with a pre-trained epoch-wise global feature map submodel followed by a sequence-to-sequence RNN;
IITNet~\cite{SEO2020102037}, the source of our 31 folds, extracting multiple features per epoch through CNNs and comparing them through sequence-wise RNNs;
GraphSleepNet~\cite{jia2020graphsleepnet}, expliciting epoch-wise functional connectivity through graph learning;
Dequidt et al.~\cite{paul}, utilizing a single-step pretrained visual CNN, who both maximize MF1 performance and rebalance training sets;
and Seraphim et al.~\cite{CAIP_article}, with a similar approach to ours, though utilizing an alternative whitening (Section~\ref{ssec:preprocessing}) and {lacking in structural preservation (cf. line 4 of Table~\ref{tab:results}).}
These models were re-trained using our methodology - except for oversampling in DeepSleepNet's sequence-to-sequence submodel - {using} their published hyperparameters.
Finally, as test sets vary between models due to recording-wise border effects, we trim test set borders to enforce uniformity.
These results, averaged over all folds, are displayed in lines 5 to 9 of Table~\ref{tab:results}.
As we can see, SPDTransNet outperforms all tested State-of-the-Art models, though our lead on Dequidt et al. is minor.

Furthermore, comparing our best results (line 2) to those of lines 4 and 9 indicate that the structural preservation of our SP-MHA improves our model's performance, with or without the influence of our new whitening (Section~\ref{ssec:preprocessing}).

\section{Conclusion}
\label{sec:conclusion}

We presented SP-MHA, a novel, structure-preserving Multihead Attention bloc, and integrated it into our SPDTransNet model, designed to analyze SPD matrix sequences. We proved said model's capabilities through automatic EEG sleep staging, obtaining a high level of per-stage performance relative to the literature.
Beyond this two-step analysis, SPDTransNet can be easily adapted to a variety of problems, for instance by using only a single encoder step and/or implementing a sequence-to-sequence classification scheme.

\bibliographystyle{IEEEtran}
\bibliography{biblio}

\end{document}